\documentclass[a4paper]{svproc}
\pdfoutput=1
\usepackage{url}

\usepackage{graphicx}
\usepackage{subcaption}
\usepackage{hyperref}
\usepackage{amsmath}
\usepackage{amssymb}
\usepackage{cite}
\usepackage{bm}
\usepackage{xcolor}
\usepackage{wrapfig}
\usepackage{cleveref}
\usepackage{acro}
\usepackage{siunitx}
\usepackage[font=small,labelfont=bf]{caption}

\DeclareAcronym{OMAV}{short = OMAV, long = Omnidirectional Micro Aerial Vehicle}
\DeclareAcronym{RL}{short = RL, long = Reinforcement Learning}
\DeclareAcronym{COM}{short = CoM, long = Center of Mass}
\crefname{figure}{Fig.}{Figs.}
\crefname{section}{Section}{Sections}

\begin{document}

\mainmatter              
\title{Learning to Fly Omnidirectional Micro Aerial Vehicles with an End-To-End Control Network}
\titlerunning{Learning to Fly Omnidirectional Micro Aerial Vehicles}
\author{Eugenio Cuniato \and Olov Andersson \and Helen Oleynikova \and Roland Siegwart \and Michael Pantic}
\authorrunning{E. Cuniato et al.} 
%
\tocauthor{Eugenio Cuniato, Olov Andersson, Helen Oleynikova, Roland Siegwart, Michael Pantic}
\institute{Autonomous Systems Lab, ETH Z\"{u}rich, Switzerland\\Corresponding author: \email{ecuniato@ethz.ch}}

\maketitle              

\begin{abstract}
Overactuated tilt-rotor platforms offer many advantages over traditional fixed-arm drones, allowing the decoupling of the applied force from the attitude of the robot. This expands their application areas to aerial interaction and manipulation, and allows them to overcome disturbances such as from ground or wall effects by exploiting the additional degrees of freedom available to their controllers.
However, the overactuation also complicates the control problem, especially if the motors that tilt the arms have slower dynamics than those spinning the propellers.

Instead of building a complex model-based controller that takes all of these subtleties into account, we attempt to learn an end-to-end pose controller using \acf{RL}, and show its superior behavior in the presence of inertial and force disturbances compared to a state-of-the-art traditional controller.
\end{abstract}

\section{Introduction}
\subsection*{Motivation}
Traditionally, aerial robots have focused on flying in free space and observing rather than interacting with the environment.
Overactuated \textit{tiltrotor} platforms decouple the motion of the robot from its attitude by independently rotating their propellers arms, opening up many applications in flying both
\begin{wrapfigure}{r}{0.5\textwidth}
    \vspace{-12pt}  
    \centering
    \includegraphics[width=\linewidth]{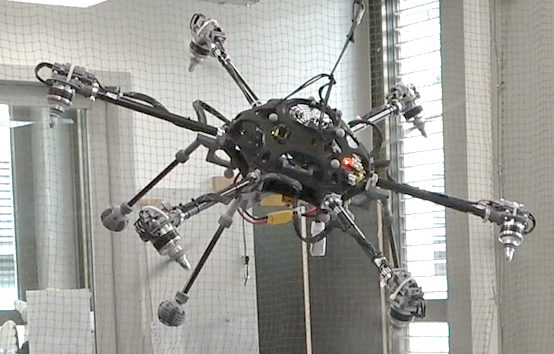}
    \caption{The \ac{OMAV} hovering at an orientation with high pitch.}
    \vspace{-25pt}
    \label{fig:intro_image}
\end{wrapfigure} close to and in contact with structure~\cite{ollero2021past}.
This particularly impacted industrial inspection and maintenance, where up-close and contact sensing are essential requirements~\cite{bodie2020active,trujillo2019novel,alexis2016aerial}.
However overactuation makes control of these platforms challenging, and flying close to or in contact with structure adds additional external disturbances such as wall effects or interaction forces.
Additionally, while propellers generally have very fast response to changes in commanded velocities, the motors that rotate the tilt arms have much slower dynamics, adding complexity to the controller model~\cite{liu2017control}. In this paper, we present an end-to-end learning approach for the pose control problem of an \acf{OMAV}. We are interested in the emergent behaviour of such a controller -- how it exploits the redundancy and variety of its actuators and how it compares to a classical controller, experimentally validating our approach on a real tiltrotor system (\cref{fig:intro_image}).
\subsection*{Problem Statement}
We investigate how an end-to-end \ac{RL} pose control policy can be learned for an overactuated aerial robot. By subjecting the agent to external disturbances and a variety of actuator model parameters during training, we learn a control policy able to exploit its different motor dynamics while behaving well in flight scenarios with changes in inertia or aerodynamic disturbances, without requiring any notably difficult modeling or additional disturbance observers.
\subsection*{Related Work}
While for traditional drones, heavily model-based control approaches have been shown to outperform PID, this has not been the case for tilt-rotors.
Several works have tried to improve upon this with, for example, Model Predictive Control (MPC)~\cite{brunner2020trajectory} or Linear Quadratic Regulators with Integral action (LQRI)~\cite{allenspach2020design}, but equal or even worse pose tracking performance than PID were observed. We believe the inherent model uncertainties of our system to be the cause of these poor performance. For this reason we investigate \ac{RL} and Domain Randomization, already used to obtain high flight performance with simple quadrotors~\cite{koch2019reinforcement,foehn2022alphapilot}, and more recently to enhance the interaction capabilities of tiltrotor aerial robots~\cite{zhang2022learning}.
\subsection*{Contributions}
We show that it is possible to learn a pose control policy able to directly command the different actuators of tiltrotor systems. We compare it to a state-of-the-art model-based controller from~\cite{bodie2020active} and highlight the interesting choices of the policy in exploiting the actuators' redundancy. Finally, we show the policy's flight performance, even in the presence of ground effects and unknown inertia disturbances.
\section{Technical Approach}
\begin{figure}[t]
    \centering
    \includegraphics{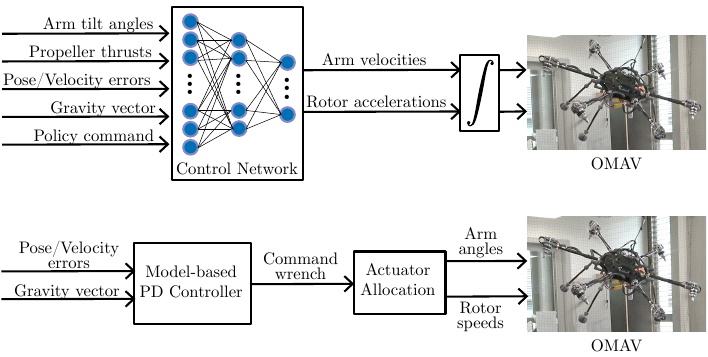}
    \caption{Proposed control network control scheme (top) and compared model-based solution from~\cite{bodie2020active} (bottom). While the model based solution divides the pose and the actuators control in two different blocks, the network learns everything end-to-end.}
    \label{fig:control_scheme}
\end{figure}
\subsection{Training environment}
We design a training environment in the GPU-accelerated NVIDIA Isaac Sim. The \ac{OMAV} flying robot is composed of i) a main body containing most of the system's inertia, ii) six rotating arms attached to the body, and iii) one propeller attached to each of the arms. The actuator dynamics are simulated by second order models, empirically tuned to approximate the real system and randomized in simulation to increase the policy's robustness. The policy runs at 100Hz.
\subsection{Neural network design}
The network is a Multi-Layer Perceptron (MLP) with three hidden layers of 256, 128 and 64 neurons respectively, using Exponential Linear Unit (ELU) activation functions, with linear activations at the output. The input observation vector contains $51$ elements: the sine and cosine of the arm tilt angles (12), the thrusts generated by the propellers (6), the full 6-degrees-of-freedom pose and velocity errors (12) of the platform as well as the previous velocity errors (6), the gravity vector projected into the current body attitude (3) and one previous network command (12). Inspired by~\cite{allenspach2020design}, the policy's output vector contains $12$ elements: six arm angular velocity commands, and six propeller acceleration commands. The network output is clipped between $[-1;1]$ and then scaled between $\pm6$ rad/s and $\pm10000$ rpm/s for the arms and propellers, respectively. 
Notice that unlike classical approaches in which the control variables are tiltrotor angles and propeller speeds directly, here we control their derivatives. This effectively commands a desired jerk on the platform, rather than a wrench. Our intuition behind this choice is twofold: on one hand, we believe this can lead to a better exploitation of the actuator dynamics, while on the other this results in smoother actuator control, as the output of the network needs to be integrated before being sent to the hardware. A comparison between the proposed learned control scheme and a state-of-the-art model based solution is shown in~\cref{fig:control_scheme}.
\subsection{Loss definition}
\label{sec:loss}
The agent trains to minimize a loss composed of three main elements.
\begin{itemize}
    \item \textbf{Velocity Loss}: $L_v = \lVert \mathbf{v} - \mathbf{v}^r \rVert + \lVert \boldsymbol{\omega} - \boldsymbol{\omega}^r \rVert $, where $\mathbf{v},\boldsymbol{\omega} \in \mathbb{R}^3$ are the current body linear and angular velocities, and $\mathbf{v}^r,\boldsymbol{\omega}^r$ are their references. The reference velocities are computed proportional to the position or attitude errors (to follow the pose reference), and limited with a $tanh$ function to limit responses to large errors. Specifically, we compute
    \begin{align}
    \label{eq:velocity_reference}
    \begin{split}
     \mathbf{v}^r &= \text{tanh}\left( k_p \mathbf{e}_p \right) , \\
     \boldsymbol{\omega}^r &= \text{tanh}\left( k_R \mathbf{e}_R \right)  ,
    \end{split}
    \end{align}
    where $k_p, k_R \in \mathbb{R}$ represent velocity gains, and the $\mathbf{e}_p, \mathbf{e}_R \in \mathbb{R}^3$ are the geometric position and orientation errors, respectively, as in~\cite{lee2010geometric}.
    \item \textbf{Posture Loss}: Given the overactuation of the system, we guide the policy to orient its rotor arms in an energy efficient way (i.e., with the arms aligned towards the gravity direction). To do that, we compute the optimally efficient pose using the model-based allocation method of~\cite{bodie2020towards}. We then penalize the distance of each arm's angle $\alpha_i \in \mathbb{R}$ from the optimal reference angle $\alpha_i^r$ using $L_p = \lVert \alpha_i - \alpha^r_i \rVert$, with $i = [1,...,6]$.
    \item \textbf{Actuation Loss}: A distribution-matching loss against the action distribution of the model-based controller. The idea is that this will guide the policy optimization towards good actions. We collected flight data from the model-based controller and recorded the actuation commands. We then fitted two Gaussian distributions to this: $\mathcal{N}_t, \mathcal{N}_p$ with means $\mu_t, \mu_p$ and variances $\sigma_t, \sigma_p$ for the tilt-arm angular velocities and propeller accelerations commands, respectively. Policy actions that do not match this distribution can then be penalized with a term inspired by the Kullback-Leibler divergence $D_{KL}(\mathcal{N}_*\,|| \mathcal{N}_\pi)$, where $\mathcal{N}_\pi$ is the policy distribution. 
\end{itemize}
\section{Experiments}
After successfully training and evaluating the controller network in simulation, we ran a variation of trained policies on the real system. We needed about 12 real world flight-tests and re-training with adjusted losses, observation noises and actuator delays until we achieved stable flight. These initial tests were also used for an evaluation of flight performance.
\subsection*{Experimental setup}To verify the real-world applicability of the in-simulation trained policies, we ran them without further modification or onboard training on the real hardware. We used a Vicon motion capture system fused with an onboard IMU (ADIS16448) to provide state estimation, and an efficient low-latency hardware interface to communicate with the arm servomotors and propeller speed controllers.
Flight performance such as pose errors and dynamics were recorded in three pose regulation experiments:
\begin{itemize}
    \item Hovering at a few centimeters above ground, to evaluate robustness against ground effects.
    \item Hovering at higher altitude, with very low aerodynamic disturbances
    \item Hovering at higher altitude but rolling $\ang{60}$, to evaluate robustness to \ac{COM} and other modeling mismatches.
\end{itemize}
\begin{wrapfigure}{r}{0.4\textwidth}
    \vspace{-24pt}  
    \centering
    \includegraphics[trim={1cm 0 1cm 0},clip,width=\linewidth]{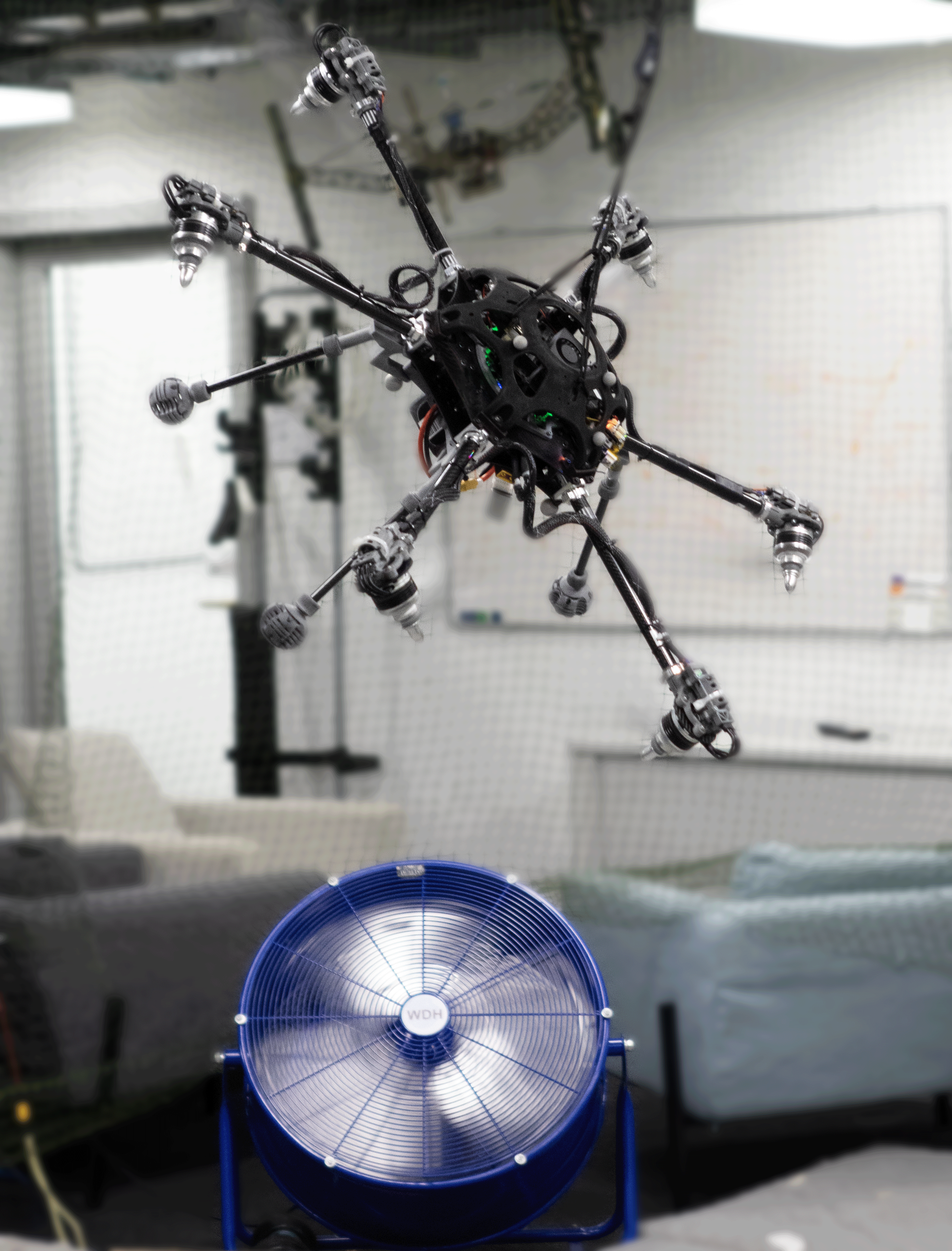}
    \caption{The \ac{OMAV} hovering with external wind disturbances.}
    \vspace{-20pt}
    \label{fig:omav_flying}
\end{wrapfigure}
The three experiments were repeated with and without a $500~\unit{g}$ weight attached to one of the \ac{OMAV}'s legs, about $10~\unit{cm}$ away from its nominal \ac{COM}. This way we aimed to evaluate the policy's robustness in different configurations against inertia and \ac{COM} changes. In particular, we compare against the state-of-the-art model based PD controller presented in~\cite{bodie2020active}, evaluating different metrics:
\begin{itemize}
    \item How the learned controller differs in its use of the redundant thrust vectoring.
    \item How is the stability and tracking quality with respect to modeling mismatches such as mass, mass distribution and system inertia.
    \item How are un-modeled external wind disturbances absorbed.
\end{itemize}
\section{Results}
\subsection{Training}
The policy is trained inside the NVIDIA Isaac Simulator with $8192$ parallel
\begin{wrapfigure}{r}{0.47\textwidth}
    \vspace{-20pt}  
    \centering
    \includegraphics[width=\linewidth]{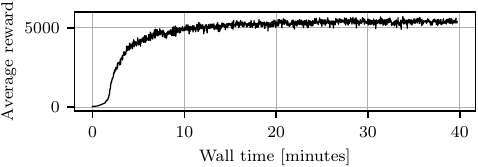}
    \caption{The training reward over time. The policy learns how to hover in only $3$ minutes, mostly converging after $10$. The rest of the time is used for fine tuning.}
    \vspace{-10pt}
    \label{fig:reward}
\end{wrapfigure}
environments, taking a total of $40$ minutes and $750$ million simulator interactions on a RTX4090 GPU, which corresponds to around 87 days of continuous flight in the real world.
Interestingly, the policy learns to hover in only $\sim$3 minutes, using the rest of the time to improve the velocity tracking performance and actuator usage.
We noticed that, even if the policy achieved most of the reward in the first $10$ minutes of training, it would use its actuators quite aggressively. The actuation loss in~\cref{sec:loss} takes care of smoothing this response. However, its relative weight in the total loss needs to be kept low in order to avoid the policy choosing an overly conservative behavior at the beginning of the training. For this reason we wait until $40$ minutes before stopping the training, after this fine-tuning phase of the actuators usage. We believe that it would be possible to shorten the total training time by better balancing the relative weights between the velocity and the actuation losses.
\subsection{Real world flight tests}
\begin{figure}
    \centering
    \includegraphics[width=\linewidth]{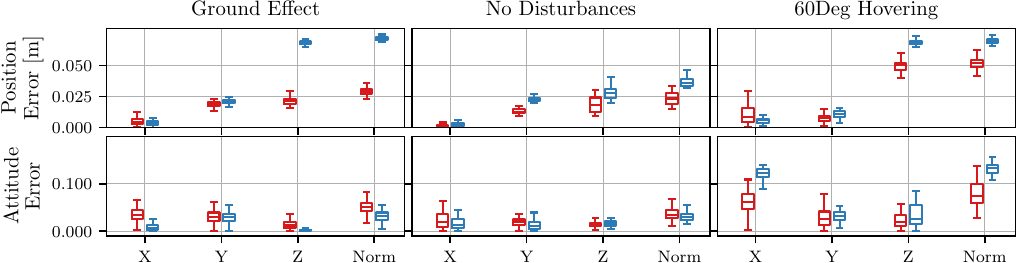}
    \caption{Pose errors norms with the nominal system. The network (red) and classic (blue) controllers are depicted along the $x$, $y$, $z$ axes and in one total norm. While the attitude errors are similar, the network outperforms the model-based solution especially in the vertical position error, probably due to a mass mismatch.}
    \vspace{-20pt}
    \label{fig:free_flight}
\end{figure}

\begin{figure}
    \centering
    \includegraphics[width=\linewidth]{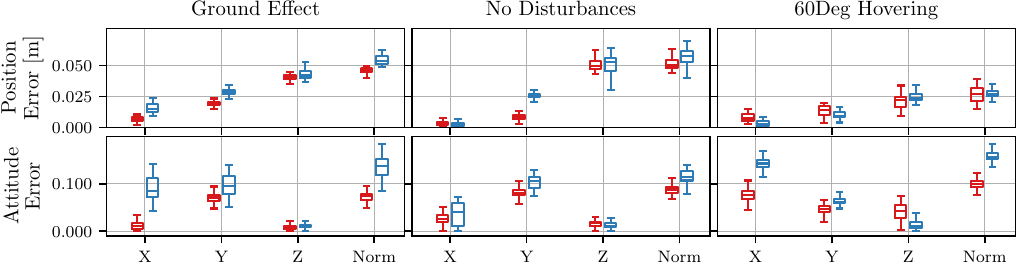}
    \caption{Pose errors norms with a $500~\unit{g}$ weight attached to one of the \ac{OMAV}'s legs. The network (red) controller now has a worse position response than in \cref{fig:free_flight}, but similar to the classic (blue) controller. On the other hand, the network attitude response outperforms the classic controller. }
    \vspace{-20pt}
    \label{fig:weight}
\end{figure}

\begin{figure}
    \centering
    \includegraphics[width=\linewidth]{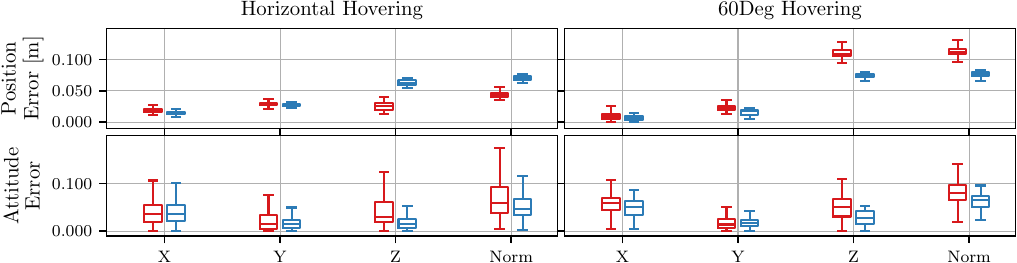}
    \caption{Pose errors norms with additional wind disturbances. The network (red) is comparable to the classic (blue) controller while horizontal, but degrades both in position and orientation control while tilting, due to the wind directly blowing into the propellers.}
    \label{fig:aerial_disturbance}
\end{figure}

To evaluate the policy's performance, we fly the \ac{OMAV} in three different pose configurations. \cref{fig:free_flight} shows the results without any additional mass, and \cref{fig:weight} with a  disturbance mass of $500~\unit{g}$ attached to one of the \ac{OMAV}'s legs. Interestingly, the policy generally exhibits better or comparable position and attitude error norms. Even without the additional mass, the model-based controller struggles because of inherent model mismatches of the platform. Notably, it outperforms the altitude tracking of the classical controller in ground effects and the orientation tracking when hovering at a $\ang{60}$ roll angle. This holds even with the additional mass and \ac{COM} mismatch.
We also compare the performance of the network in two additional scenarios: hovering far from the ground and while at $\ang{60}$ roll angle, without the disturbance mass but with additional wind disturbances.
The pose error results are in \cref{fig:aerial_disturbance}. While the network position error is lower during the horizontal hovering, the model-based controller outperforms the network while rolling, especially in the vertical axis. The results on the attitude control are comparable with a slight advantage of the model-based controller, being less oscillatory than the policy. 
The fan blowing directly into the propellers creates a much more rapidly changing disturbance than the air cushion generated by the ground effect. Since we train the policy only with slow changing disturbances (random forces and torques applied every $3$ seconds), the controller doesn't effectively know how to counteract it. To improve the policy's response, time-varying single propeller disturbances could be added in simulation, to recreate a chaotic airflow.
\section{Main Experimental Insights}
\subsection{Model-based vs. Network actuators usage}
\begin{figure}
    \centering
    \includegraphics[width=\linewidth]{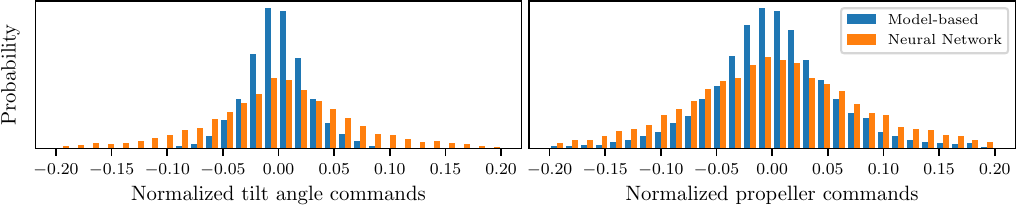}
    \caption{Control signals distributions comparison. The actuator commands are normalized in the range $[-1;1]$, but both the policy and the model-based controller effectively utilize up to 20\% of this bandwidth to control the platform. }
    \label{fig:actuator_distributions}
\end{figure}

We utilize the Actuator Loss in \cref{sec:loss} to guide the network optimization towards a working local minima. However, we keep the weight of that loss relatively low, allowing the policy to explore more efficient or robust solutions. For this reason, 
as in~\Cref{fig:actuator_distributions}, the learned network learns to use the tilting arms much more than the model-base controller. While the propellers command distribution is wider but similar to the model-based one, the tilt rotors are much more actuated by the policy, with a much wider distribution.
\subsection{Thrust geometry}
One interesting emerging behaviour of the policy is its choice of thrust geometry, i.e. it's choice of tilt angles redundancy while flying. The model-based approach tries to keep the angles always oriented along gravity, such to spend most of the energy to counteract gravity and reducing the amount of wasted internal forces.
\begin{wrapfigure}{r}{0.5\textwidth}
    \vspace{-20pt}  
    \centering
    \includegraphics[width=\linewidth]{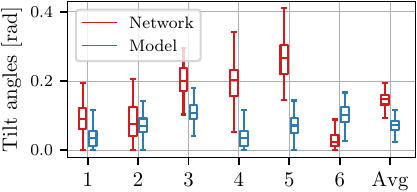}
    \caption{Angle commands for the six rotor arms and their average. The network (red) actuates the tilt-arms more often while flying (higher variance), and keeps them more tilted (higher mean) w.r.t. the model-based solution (blue).}
    \vspace{-20pt}
    \label{fig:tilt_angles}
\end{wrapfigure}
On the other hand, even if we give the policy an additional reward to follow the same approach, it actually tends to increase the tilt angles generating more internal forces. For example, we can observe this behavior from the recorded arm angular command of the hovering experiment in~\Cref{fig:tilt_angles}.
This behaviour probably arises from the constant external wrench disturbances that affect the simulated \ac{OMAV} while flying. 
Indeed, more internal forces make the system more reactive in counteracting disturbances in the lateral $x$ and $y$ body directions, especially around the horizontal hovering configuration.
For example, counteracting lateral forces while hovering in a minimum energy configuration (with the propellers point down along gravity's direction) would require the arms to first tilt in the disturbance's direction, and then the propellers to increase their speed to push against it. However, if the arms are already partially tilted, then this only requires increasing the propellers speeds, which is faster than also rotating the arms.

\section{Conclusions}
Our work has demonstrated that an end-to-end controller can be learned for a complex overactuated aerial robot dealing with large external disturbances or model mismatches. Importantly, the policy trains on a distribution of hypothetical robots in simulation - but then runs without re-tuning on a specific instance of robot with all its inherent physical properties and imperfections.
Our policy learns to reject any disturbances seen in simulation, and performs better than a classical controller in the presence of ground effects and mass disturbances. These are common scenarios experienced in aerial manipulation tasks.
However, as we show in the high wind scenario, the policy cannot reject fast-varying aerodynamic disturbances that it did not see during training. 
Future work will investigate learning only single elements of the control pipeline (the pose controller or the actuator allocation) and comparing it with the end-to-end system presented there. Also, we believe that additional simulated aerodynamic disturbances could improve the system's wind response.

\section{Acknowledgments}
This work has been supported by the European Unions Horizon 2020 Research and Innovation Programme AERO-TRAIN under Grant Agreement No. 953454 and the ETH RobotX Research Program.
%
%
%
\bibliographystyle{ieeetr}
\bibliography{references}
\end{document}